\documentclass[conference]{IEEEtran}
\usepackage{blindtext, graphicx}
\ifCLASSINFOpdf
\else
\fi
\usepackage{url}
\usepackage{xcolor}
\usepackage{graphicx}

\graphicspath{{figs/}}
\begin{document}
\title{Decentralized Control of a Hexapod Robot Using a Wireless Time Synchronized Network}

% \author{\IEEEauthorblockN{Michael Shell}
% \IEEEauthorblockA{School of Electrical and\\Computer Engineering\\
% Georgia Institute of Technology\\
% Atlanta, Georgia 30332--0250\\
% Email: http://www.michaelshell.org/contact.html}
% \and
% \IEEEauthorblockN{Homer Simpson}
% \IEEEauthorblockA{Twentieth Century Fox\\
% Email: homer@thesimpsons.com}
% \and
% \IEEEauthorblockN{James Kirk\\ and Montgomery Scott}
% \IEEEauthorblockA{Starfleet Academy\\
% San Francisco, California 96678-2391\\
% Telephone: (800) 555--1212\\


\author{\IEEEauthorblockN{James Fang,
Dinesh Parimi, 
Arjun Dhindsa,
Craig B. Schindler, and
Kristofer S. J. Pister}
\IEEEauthorblockA{Berkeley Sensor~\&~Actuator Center\\
Department of Electrical Engineering and Computer Sciences\\
University of California, Berkeley\\
Berkeley, California, USA\\
Email: \{jamesfang,
dineshp,
arjun.dhindsa,
craig.schindler,
ksjp\}@berkeley.edu
}}
% \IEEEauthorblockA{\IEEEauthorrefmark{2}Twentieth Century Fox, Springfield, USA\\
% Email: homer@thesimpsons.com}
% \IEEEauthorblockA{\IEEEauthorrefmark{3}Starfleet Academy, San Francisco, California 96678-2391\\
% Telephone: (800) 555--1212, Fax: (888) 555--1212}
% \IEEEauthorblockA{\IEEEauthorrefmark{4}Tyrell Inc., 123 Replicant Street, Los Angeles, California 90210--4321}}


\maketitle

\begin{abstract}
%\boldmath
Robots and control systems rely upon precise timing of sensors and actuators in order to operate intelligently.
We present a functioning hexapod robot that walks with a dual tripod gait; each tripod is actuated using its own local controller running on a separate wireless node.
We compare and report the results of operating the robot using two different decentralized control schemes.
With the first scheme, each controller relies on its own local clock to generate control signals for the tripod it controls.
With the second scheme, each controller relies on a variable that is local to itself but that is necessarily the same across controllers as a by-product of their host nodes being part of a time synchronized IEEE802.15.4e network.
The gait synchronization error (time difference between what both controllers believe is the start of the gait period) grows linearly when the controllers use their local clocks, but remains bounded to within 112 microseconds when the controllers use their nodes' time synchronized local variable.
\end{abstract}

% \begin{IEEEkeywords}
% IEEEtran, journal, \LaTeX, paper, template.
% \end{IEEEkeywords}

% \fi
\IEEEpeerreviewmaketitle

\section{Introduction}

Control systems can be fully centralized, fully decentralized, or a combination of both.
In a centralized control system, all of the information about the system as well as the calculations based on that information take place at a single location.
In a decentralized system, the information about the system as well as the calculations about that information take place at multiple locations~\cite{sandell1978survey}.

Decentralized control is desirable for a multitude of reasons. Firstly, by modularizing a robotic system, lengthy control wires running from the central controller to various different parts can be cut down, leaving just the power wires needed to supply current. Wiring is one of the most vulnerable points on any robot and is thus one of the most common sources of faults and breaks~\cite{flynn1989building}. By reducing the wiring required, the robustness of the entire system is improved. Additionally for microrobots, having lots of wires is far too costly in size and weight and can easily make micro-scale systems infeasible~\cite{contreras2016durability}.

Secondly, decentralized control allows for more flexible and responsive robotic systems~\cite{flynn1989building}. This becomes especially apparent in robot swarms~\cite{bergbreiter2003cotsbots} and self-assembling robots~\cite{rubenstein2014programmable} where the components of the system may not be physically connected in any way.

\begin{figure}
\includegraphics[width=\linewidth]{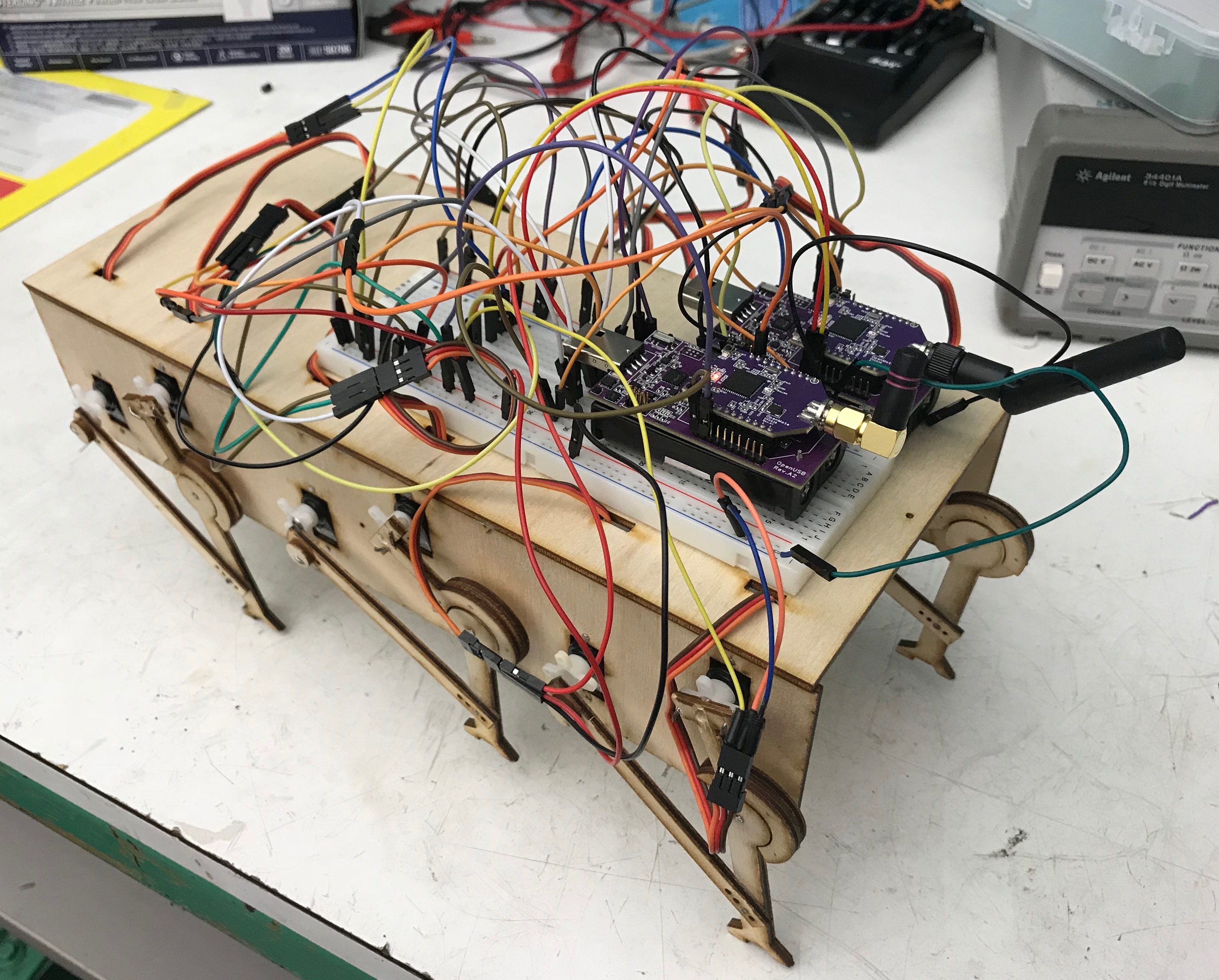}
\caption{Larry the hexapod with two on-bot OpenMotes sitting on his frame patiently awaits commands from the central controller or "DAGroot" mote (not shown), which is connected via OpenUSB to a computer and controlled by a human driver using a python GUI.}
\label{hexapod}
\end{figure}

One of the biggest challenges to decentralized control is time synchronization of the various distributed components. In this paper, we demonstrate the viability of implementing the low-power wireless protocol stack OpenWSN~\cite{watteyne2012openwsn} on OpenMotes~\cite{vilajosana2015openmote} for the time synchronization of decentralized robotic movement on Larry~(Fig.~\ref{hexapod}), a hexapod robot that uses two independent but wirelessly connected OpenMotes as controllers for its legs.

\section{Time Synchronized Protocol}

OpenWSN~\cite{watteyne2012openwsn} and Contiki~\cite{dunkels2004contiki} provide open-source implementations of a low-power, high-reliability
communication stack based on the IEEE 802.15.4e~\cite{ieee802154e} protocol and the IETF
6LoWPAN~\cite{6lowpan} and 6TiSCH~\cite{6tisch} protocols.
IEEE802.15.4e added Timeslotted Channel Hopping (TSCH) to the IEEE802.15.4 standard, allowing for ultra low power communication with very high reliability. 6LoWPAN allows IPv6 packets to be sent over lossy and low-power wireless sensor networks, allowing them to interface directly with the Internet, and 6TiSCH provides standard ways of implementing and managing a TSCH schedule. CoAP~\cite{shelby2014constrained} is a web transfer protocol specifically designed for constrained and low-power networks.
Using IEEE 802.15.4e TSCH with centralized feedback control has been demonstrated~\cite{schindler2017implementation}; however, the work in this paper uses IEEE802.15.4e TSCH for decentralized control by exploiting the extremely tight time-synchronization of nodes.
Very small synchronization errors between 3-hop deep nodes in TSCH networks have been demonstrated using Contiki (2 microseconds)~\cite{elsts2016microsecond}, OpenWSN (76 microseconds)~\cite{chang2015adaptive}, and Dust Networks' SmartMesh software and LTC5800 chip (5 microseconds)~\cite{ltc5800}.

\begin{figure}
\includegraphics[width=\linewidth]{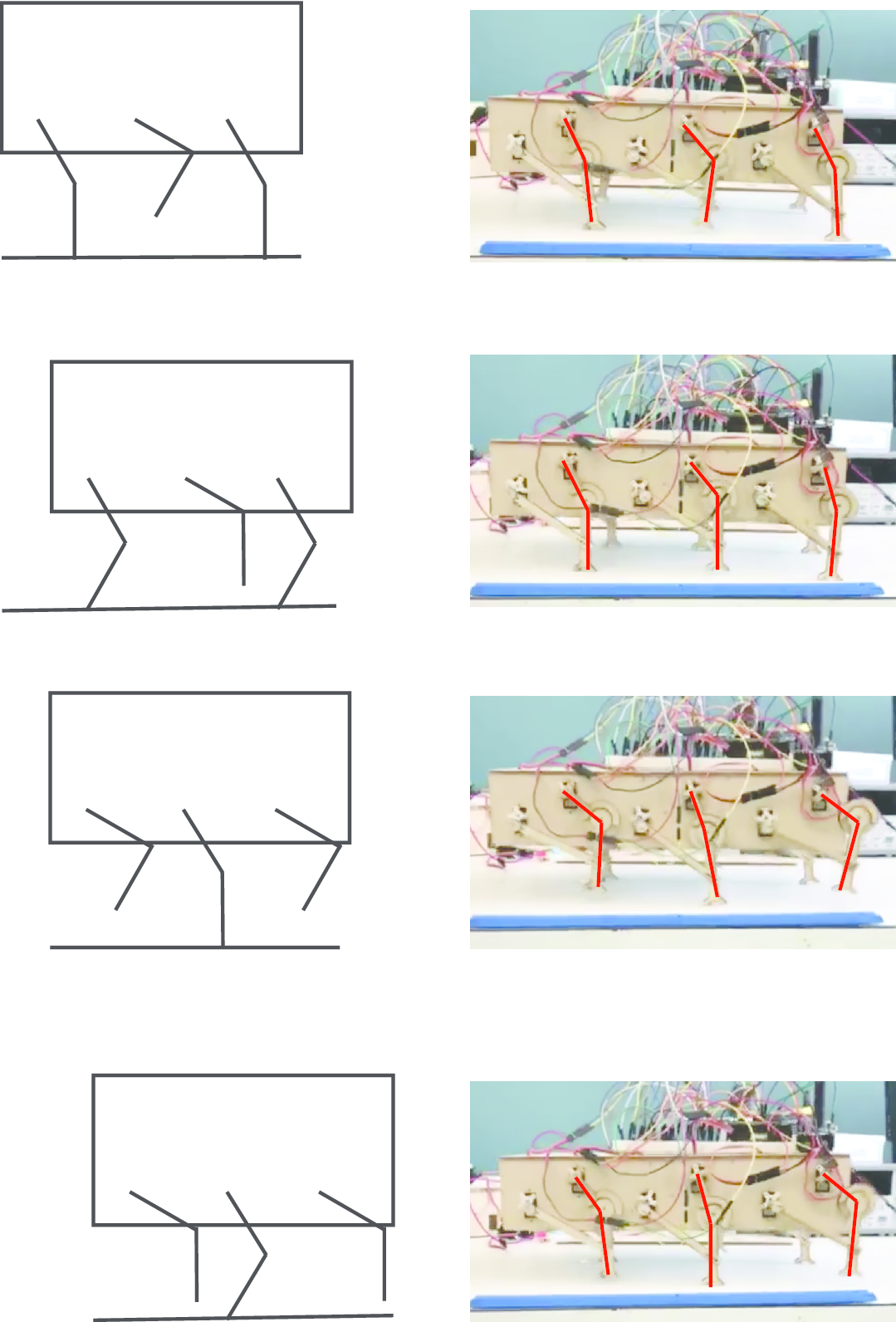}
\caption{The dual tripod gait in action (side view, walking to the right). (i) The right tripod lowers and the left tripod lifts at the hip. (ii) The right tripod propels Larry forward by rotating back at the knee while the left tripod rotates forward at the knee to reset. (iii) The right tripod lifts, and the left tripod lowers at the hip. (iv) The left tripod propels Larry forward by rotating back at the knee while the right tripod rotates forward at the knee to reset. These four actions loop to produce the forward-moving dual tripod gait.}
\label{dual_tripod_gait}
\end{figure}

\begin{figure}
\includegraphics[width=\linewidth]{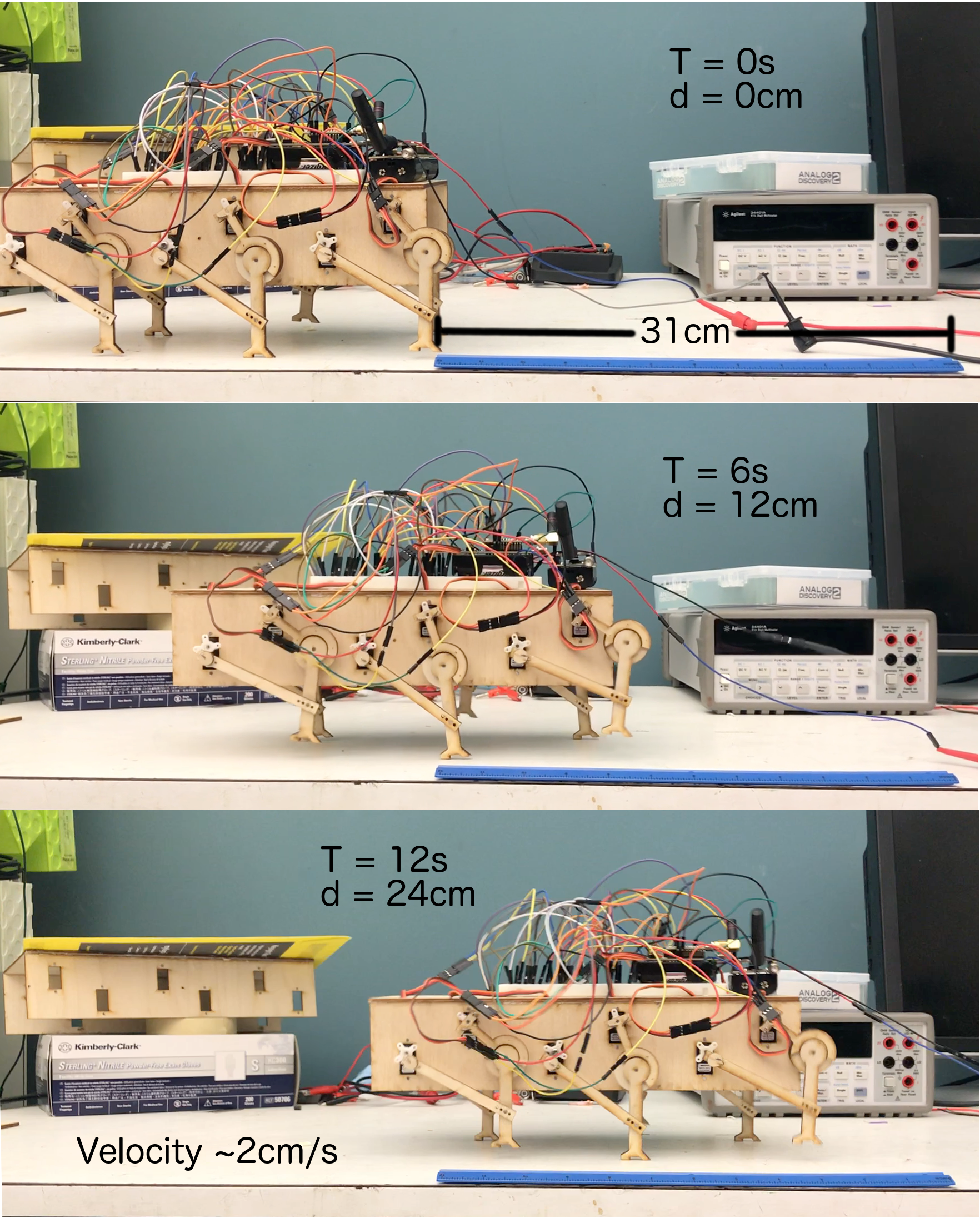}
\caption{An image panel of Larry walking and being controlled by $M_1$ and $M_2$ using scheme $S_2:$ decentralized synchronized control (ASN synchronization). Larry's approximate velocity is around 2 cm/s.}
\label{velocity_figure}
\end{figure}

\section{Larry, an example platform}
\subsection{Design and Construction}
Larry, being a hexapod, has six legs, each with two points of rotation (i.e. joints), which are referred to as the ``hip" and ``knee". These joints are each actuated by a servo motor for a total of twelve servos. \par
The hip-and-knee design allows for two rotational degrees of movement: the servo actuating the hip translates its rotational motion into linear vertical motion, allowing Larry to lift himself, while the servo actuating the knee translates to linear horizontal movement, allowing Larry to pull himself forwards or backwards. The advantage of this design is simplicity: there are only three moving parts on the leg. Legs also allow Larry to engage in step-like motions, which can be adapted to traverse over obstacles in the human world (i.e. stairs, gaps, etc.). \par 

The hexapod frame and legs are fabricated by laser cutting $\frac{1}{8}$ inch plywood. The parts are assembled using small nails. Six servos are mounted directly on a $30 \times 6$ cm sideboard. Two mirrored sideboards are then attached to a $30 \times 12.5$ cm top frame. The legs are attached to the servos at the appropriate joints. The wiring is done using a breadboard, which sits on the top frame with two OpenMotes.

\subsection{Gait}

Larry walks using a dual tripod gait. As the name suggests, there are two tripods, each one consisting of the front and back leg from one side of the robot and the middle leg from the other side. At any point in time, at least one tripod will be in contact with the ground while the other tripod resets in midair. The gait has four main positions which when executed in a loop results in step-like motion~(Fig.~\ref{dual_tripod_gait}). \par

A single tripod goes through the following four events: 
\begin{enumerate}
    \item The tripod actuates the hip downwards, lifting Larry. 
    \item The tripod actuates the knee backwards, propelling Larry forward.
    \item The tripod actuates the hip back upwards.
    \item The tripod actuates the knee forwards to reset for the next iteration.
\end{enumerate}
Larry can be seen walking in Fig.~\ref{velocity_figure}.
The tripods are offset by exactly 180 degrees, meaning that in our four-step gait, tripod 1 should be exactly two steps ahead of tripod 2 at all times. Since the gait motion is periodic with a period of one second, tripod 1 should be half a second ahead of tripod 2 at all times. This is where time synchronization becomes crucial for proper execution of the robotic gait. If the gait offset of the tripods is out-of-sync, Larry would not be able to walk properly. In the worse case, if tripod 1 and tripod 2 are 180 degrees out-of-sync (i.e. they are performing the same gait actions at the same time), Larry will not be walking at all but instead be squatting in place over and over. \par

\begin{figure}
\includegraphics[width=\linewidth]{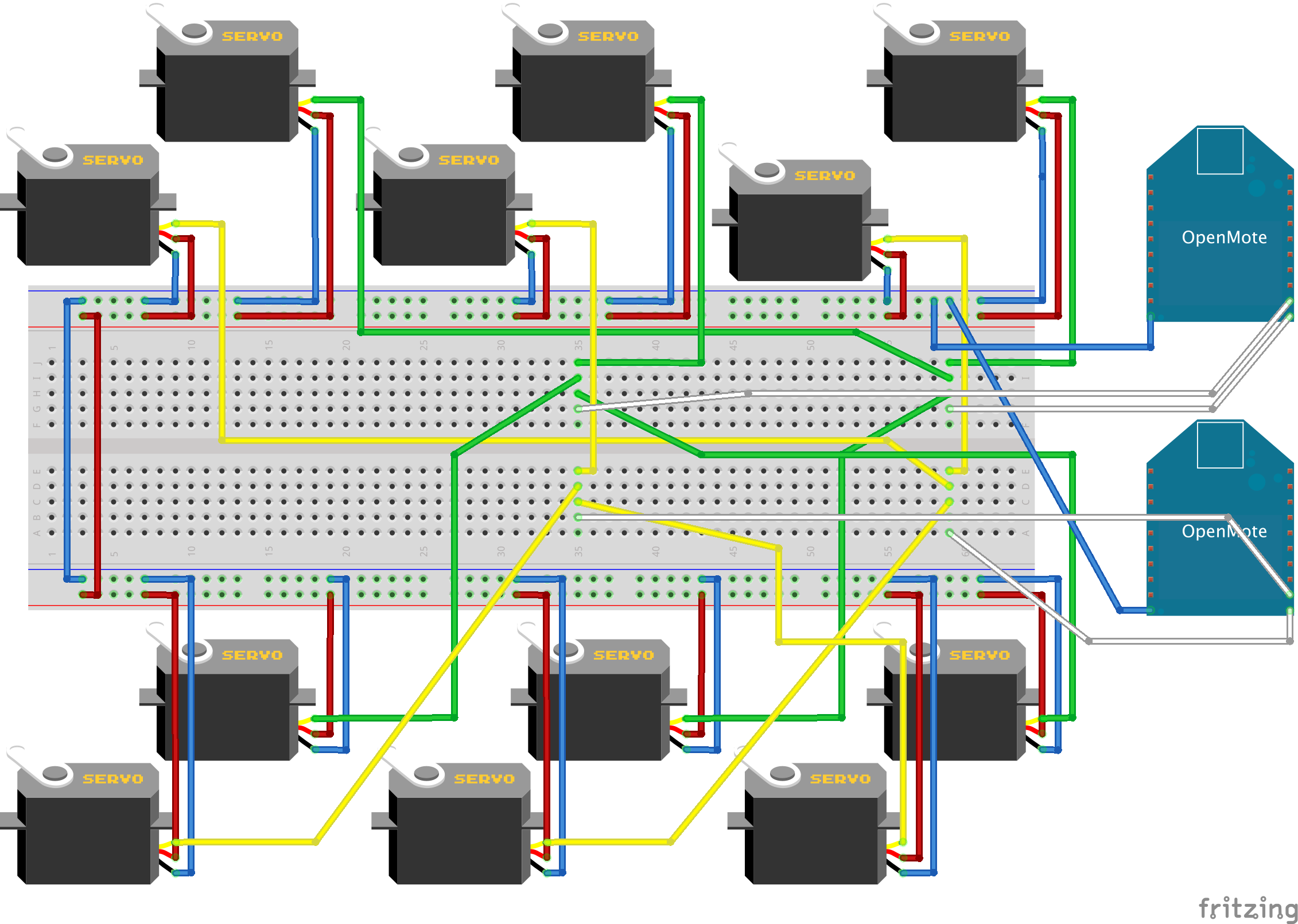}
\caption{Circuit diagram of the servos being controlled by two motes. $M_1$ (top mote) is controlling the hip servos (green control lines), and $M_2$ (bottom mote) is controlling the knee servos (yellow control lines). The servos are powered by an external supply drawing 2A at 5V.}
\label{ciruit_diagram}
\end{figure}

The gait events are controlled using two OpenMote-cc2538 boards or ``motes". Let the two motes be $M_1$ and $M_2$. $M_1$ controls all six hip servos, and $M_2$ controls all six knee servos. A circuit diagram of the two motes connected to the servos that they each control can be seen in Fig.~\ref{ciruit_diagram}. \par
%Let $T_1$ and $T_2$ be tripod 1 and tripod 2 respectively. Within a gait period: 1) $M_1$ tells $T_1$'s vertical actuators to turn "down" as $T_2$'s vertical actuators are "raised up". At this point, $T_1$'s legs are making contact with the ground, while $T_2$'s legs are suspended in mid air. 2) $M_2$ then tells $T_1$'s horizontal actuators to turn "backwards", pulling the hexapod forwards. Meanwhile $T_2$'s horizontal actuators are reset to the "forward" position. 3) $M_1$ then tells $T_1$'s vertical actuators to raise up and $T_2$'s vertical actuators to turn down. 4) $M_2$ then tells $T_1$'s horizontal actuators to reset to the forward position and $T_2$'s horizontal actuators to turn backwards, pulling the hexapod forwards. This is then repeated every period (i.e. every second).% \par 

\section{Firmware and Networking Implementation}
Hexapod control is implemented with a modified version of the OpenWSN wireless protocol stack firmware, running on OpenMote-cc2538 Rev. A1 boards. OpenWSN designates one mote as a ``DAGroot" that can send and receive messages from all motes connected in the network.\par
\begin{figure}
\includegraphics[width=\linewidth]{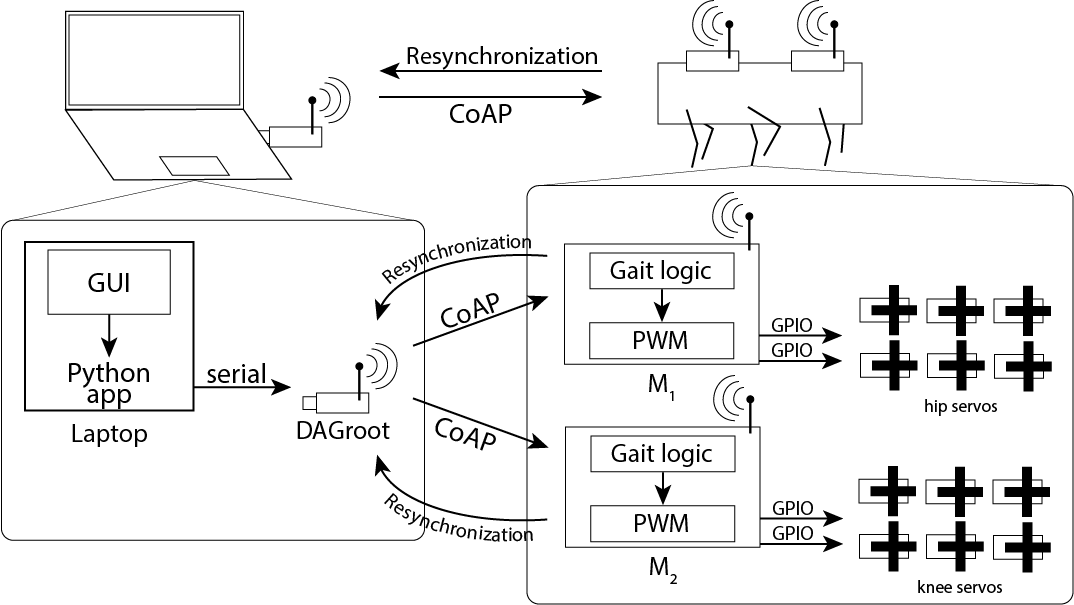}
\caption{A human driver can operate Larry using a Python GUI interface, which is processed by a Python application that then sends a constrained application protocol (CoAP) message, via the connected DAGroot, telling the child motes carried on Larry's back to start running their respective gait events which drive the servos on Larry's legs. Top: physical system of a laptop with DAGroot connected communicating with the two child motes on Larry. Bottom: block diagram of the control logic occurring within the physical systems.}
\label{control_diagram}
\end{figure}

The firmware is flashed to $M_1$, $M_2$ and the DAGroot motes. A human driver can interact using a CoAP application~(Fig.~\ref{control_diagram}), usable through either a Python GUI interface or the command line (e.g. the driver can send a CoAP message from the DAGroot to $M_1$ and $M_2$ telling them to start running their gait events).
Three separate methods for gait control were explored: centralized control, decentralized open loop control, and decentralized synchronized control. \par
In centralized control, the DAGroot was responsible for timing the gait events and sending information over the network about the angle of each servo motor (i.e. turning it) to the two child motes, whose only task was actually implementing the pulse width modulation (PWM) signals to drive the motors. However, in this implementation, the DAGroot was constantly transmitting messages to the other motes, typically several times per second, and this caused latency issues since the messages to both motes cannot be sent at exactly the same time and the transmission itself introduces some delay. \par
To work around this issue, we tried decentralized open loop control. The DAGroot was now only responsible for sending high-level commands to the child motes such as ``go forward", ``turn left", or ``stop", and the child motes used their independent, internal timing schemes to control the gait events driving the servo motors. In practice, this worked well while driving Larry for short amounts of time (on the order of minutes). However, the on-board timers of the motes have a drift of at most 10 ppm, so for longer runs, the two motes will eventually grow out-of-sync. Thus, we need to periodically resynchronize the two child motes in order to mitigate timing drift between the gait events. \par 
We addressed this issue in decentralized synchronized control by synchronizing the motes' timing schemes using the network's absolute slot number (ASN), a local variable on each mote that is incremented every 15 milliseconds according to the mote's local clock. Anytime a message is sent between the child motes and the DAGroot, the child motes will resynchronize their clocks to the DAGroot's clock - this makes the ASN a reliable timing reference across all motes. As per the default OpenWSN timing requirements, motes must be synchronized within 1 millisecond. Our implementation has the child motes communicating with the DAGroot every 30 seconds in a worst case scenario, allowing the child motes to be synchronized within 120 microseconds (see results below). This implementation allows Larry to walk for any duration of time without squatting (i.e. desynchronizing). \par

\begin{figure}
\includegraphics[width=\linewidth]{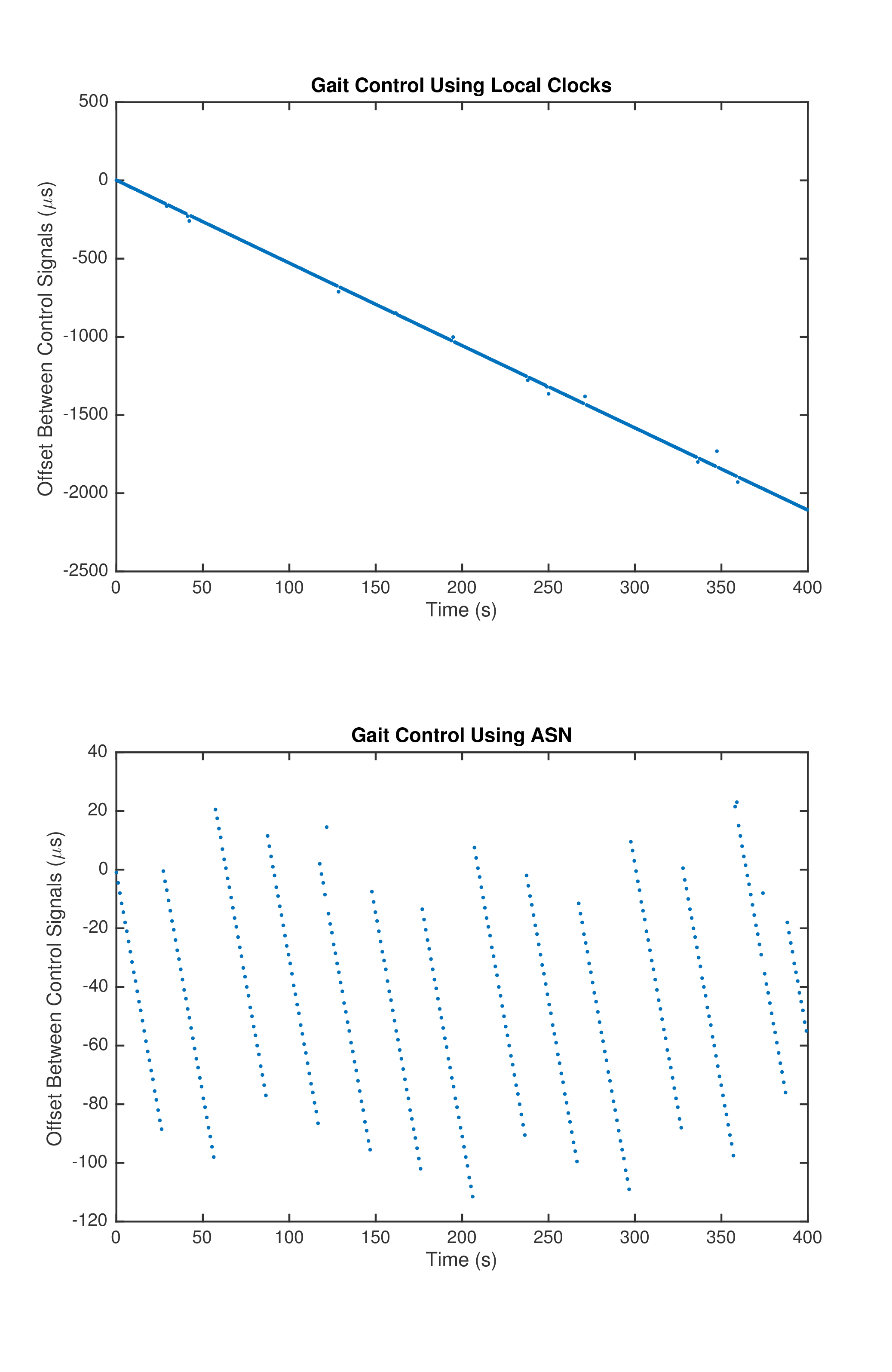}
\caption{Gait synchronization error between the two motes starting from an initial state of zero error. Top: the motes are running on decentralized open loop control using independent on-board timers and are experiencing a drift of around 5 $\mu$s/s (5 ppm) for a total of slighlty more than 2~ms over the 400~second experiment. This drift continues indefinitely. Bottom: the motes are running on decentralized synchronized control using the network's ASN and are experiencing a drift of around 3 $\mu$s/s (3 ppm). The motes resynchronize every 30 seconds, so the offset between them never exceeds 112 $\mu$s.}
\label{offset_data}
\end{figure}

\section{Results}

With decentralized control, $M_1$ and $M_2$ are  executing their  respective  gait events independently, yet the events must be time synchronized for proper gait performance. Our results compare the gait synchronization error using two gait control schemes: $S_1:$ decentralized open loop control and $S_2:$ decentralized synchronized control. Decentralized open loop control uses the free-running onboard clocks of $M_1$ and $M_2$ as independent references for time. Decentralized synchronized control relies on the ASN as a global time reference for making control decisions. If $M_1$ and $M_2$ have the same reference for the start of each gait period, their gait events will be performed in sync. The quality of the control will then be measured by the difference between the starts of $M_1$ and $M_2$'s gait period. \par

$S_1$ lets $M_1$ and $M_2$'s clocks run independently of each other, but they may have an error up to 10 ppm (as specified by the 32.768kHz crystal the OpenMote-cc2538 uses). This implies that one of the clocks will be running slightly faster than the other (less than 10 microseconds faster per second). As expected, this error is causing a drift between $M_1$ and $M_2$'s gait period starts, causing them to gradually fall out of sync over time~(Fig.~\ref{offset_data}, top). For the 400-second sample run, the gait synchronization offset increases linearly to slightly over 2 milliseconds with a slope of -5 $\mu$s/s. For a gait period of 1 second, a 2 millisecond difference does not have much noticeable effect on gait performance (i.e. $M_2$, controlling the knees, is pulling the hexapod forwards 2 milliseconds later than it should). However, after slightly less than 28 hours of operation, the drift would cause $M_1$ and $M_2$ to be 180 degrees out of phase, causing Larry to squat instead of walk. He would then be in an uncontrollable state. \par

$S_2$ uses the ASN variable on $M_1$ and $M_2$ as a time reference to make control decisions. $M_1$ and $M_2$ will resynchronize their clocks anytime communication happens with the DAGroot (their time-source parent). For the 400-second sample run~(Fig.~\ref{offset_data}, bottom), the offset between the two motes is never more than 112 microseconds, i.e. the two motes will never be out of sync by more than 112 microseconds. The slope of the line is around -3 microseconds per second, showing an error of 3 ppm between these two motes. The bounded gait synchronization error means that Larry can walk indefinitely without getting into an uncontrollable state. \par

The synchronization bound can be further reduced from 112 microseconds to 30 microseconds by having the child motes communicate with the DAGroot more often. The above data was collected using a worst case communication period of 30 seconds, resulting in a bound of 112 microseconds. Using the 32.768 kHz clock on the motes, in the best case we are limited to the clock period of $1/(32.768$ kHz$)~\approx~30$~microseconds. By simply reducing the worse case communication period to 10 seconds, we were able to achieve a synchronization bound of around 30 microseconds (the 112 microsecond-bounded data is used here for ease of visual analysis). And with more sophisticated techniques such as adaptive synchronization this bound could be reduced even further~\cite{chang2015adaptive}. Microsecond time synchronization using TSCH has been demonstrated by Elsts et al.~\cite{elsts2016microsecond}. \par

The results show that for gait synchronization between $M_1$ and $M_2$, the scheme $S_2$, taking advantage of the TSCH mode of IEEE802.15.4 with the usage of the ASN implemented by OpenWSN, has the objectively better decentralized control performance.

\section{Conclusion}
Decentralized control can be used to successfully coordinate actions in robotic systems. However, as the results show, success depends on keeping the various controllers time-synchronized. The controllers themselves cannot be relied on to stay time-synchronized due to their clock error of 10 ppm. Thus, the controllers need to be connected via a network to share a common time reference. The OpenWSN network presented here uses the TSCH mode of IEEE802.15.4 to ensure each controller's time reference remains synchronized with respect to the DAGroot's clock by usage of the ASN variable implemented by OpenWSN. \par
The application of a network-synchronized robotic system extends beyond the single-robot gait presented in this paper. Precisely coordinated decentralized control can be used in a wide range of robotic systems: for example, self-reconfiguring robots or intelligent robot swarms. Furthermore, the use of synchronized nodes implementing the OpenWSN protocol stack can be extended beyond just the three-node topology presented in this paper. There have been projects such as SensLab which uses 1024 synchronized nodes at once~\cite{des2011two} and FIT IoT-LAB, a testbed for IoT technologies which is composed of 2728 synchronized wireless nodes and 117 mobile robots~\cite{adjih2015fit}.

\section*{Acknowledgment}

The authors would like to thank the Berkeley Sensor~\&~Actuator Center and the UC Berkeley Swarm Lab for supporting this project.

\ifCLASSOPTIONcaptionsoff
  \newpage
\fi

\bibliographystyle{IEEEtran}
\bibliography{IEEEabrv,bib.bib}


%\vfill


\end{document}